\documentclass[11pt,letterpaper]{article}
\usepackage{emnlp2017}
\usepackage{times}
\usepackage{mdwlist}
\usepackage{latexsym}
\usepackage{mathtools}
\usepackage{amsfonts}
\usepackage{amssymb}
\usepackage{booktabs}
\usepackage{xcolor}
\usepackage{enumitem}
\usepackage{lipsum}
\usepackage{graphicx}
\usepackage{subcaption}
\usepackage{xspace}
\usepackage{algorithmic,algorithm}
\usepackage{bbm}

\renewcommand{\vec}[1]{\boldsymbol{#1}}

\renewcommand{\sectionautorefname}{\S\kern-0.2em}
\renewcommand{\subsectionautorefname}{\S\kern-0.2em}
\renewcommand{\subsubsectionautorefname}{\S\kern-0.2em}

\newcommand{\bleu}{\textsc{Bleu}\xspace}
\newcommand{\persentence}{Per-Sentence \bleu}
\newcommand{\heldout}{Heldout \bleu}

\newcommand{\ignore}[1]{}

\emnlpfinalcopy

\def\emnlppaperid{1355}

\title{Reinforcement Learning for Bandit Neural Machine Translation with Simulated Human Feedback}

\newcommand{\idcs}{${}^\odot$}
\newcommand{\idlsc}{${}^\spadesuit$}
\newcommand{\idumiacs}{${}^\diamondsuit$}
\newcommand{\idischool}{${}^\clubsuit$}
\newcommand{\idmsr}{${}^\heartsuit$}

\author{Khanh Nguyen\idcs\idumiacs \and Hal Daum{\'e} III\idcs\idlsc\idumiacs\idmsr \and Jordan Boyd-Graber\idcs\idlsc\idischool\idumiacs \\
  University of Maryland: Computer Science\idcs, Language Science\idlsc, iSchool\idischool, UMIACS\idumiacs\\
  Microsoft Research, New York\idmsr\\
  \tt{ \{kxnguyen,hal,jbg\}@umiacs.umd.edu }
 }

\date{}

\begin{document}

\abovedisplayskip=12pt plus 3pt minus 9pt
\abovedisplayshortskip=0pt plus 3pt
\belowdisplayskip=12pt plus 3pt minus 9pt
\belowdisplayshortskip=7pt plus 3pt minus 4pt

\maketitle

\begin{abstract}
  Machine translation is a natural candidate problem for reinforcement
learning from human feedback: users provide quick, dirty ratings on candidate translations
to guide a system to improve.  Yet, current neural machine translation training focuses
on expensive human-generated reference translations.  We describe a reinforcement learning algorithm that improves neural machine translation systems from simulated human feedback.
Our algorithm combines the advantage actor-critic algorithm~\cite{mnih2016asynchronous}
with the attention-based neural encoder-decoder architecture~\cite{luong2015effective}.
This algorithm (a) is well-designed for problems with a large action space and delayed rewards,
(b) effectively optimizes traditional corpus-level machine translation metrics, and
(c) is robust to skewed, high-variance, granular feedback modeled after actual human behaviors.

\end{abstract}

\section{Introduction}

Bandit structured prediction is the task of learning to solve complex joint prediction problems (like parsing or machine translation) under a very limited feedback model:
a system must produce a \emph{single} structured output (e.g., translation) and then the world
reveals a \emph{score} that measures how good or bad that output is, but provides neither a ``correct'' output nor feedback on any other possible output \cite{daume15lols,sokolov2015coactive}.
Because of the extreme sparsity of this feedback, a common experimental setup is that one pre-trains a good-but-not-great ``reference'' system based on whatever labeled data is available, and then seeks to improve it over time using this bandit feedback.
A common motivation for this problem setting is cost.
In the case of translation, bilingual ``experts'' can read a source sentence and a possible translation, and can much more quickly provide a rating of that translation than they can produce a full translation on their own.
Furthermore, one can often collect even less expensive ratings from ``non-experts'' who may or may not be bilingual \cite{hu2014crowdsourced}.
Breaking this reliance on expensive data could unlock previously
ignored languages and speed development of broad-coverage machine translation systems.

All work on bandit structured prediction we know makes an important simplifying assumption: the \emph{score} provided by the world is \emph{exactly} the score the system must optimize (\autoref{sec:problem}).
In the case of parsing, the score is attachment score; in the case of machine translation, the score is (sentence-level) \bleu.
While this simplifying assumption has been incredibly useful in building algorithms, it is highly unrealistic.
Any time we want to optimize a system by collecting user feedback, we must take into account:
\begin{enumerate}[noitemsep,nolistsep]
\item The metric we care about (e.g., expert ratings) may not correlate perfectly with the measure that the reference system was trained on (e.g., \bleu or log likelihood);
\item Human judgments might be more granular than traditional
  continuous metrics (e.g., thumbs up vs. thumbs down);
\item Human feedback have high \emph{variance} (e.g., different raters
  might give different responses given the same system output);
\item Human feedback might be substantially \emph{skewed} (e.g., a
  rater may think all system outputs are poor).
\end{enumerate}
Our first contribution is a strategy to
simulate expert and non-expert ratings to evaluate the robustness of bandit structured prediction algorithms in general, in a more realistic environment (\autoref{sec:noise_model}).
We construct a family of perturbations to capture three attributes: \emph{granularity}, \emph{variance}, and \emph{skew}.
We apply these perturbations on automatically generated scores to simulate noisy human ratings. 
To make our simulated ratings as realistic as possible, we study recent human evaluation data \cite{graham2017can} and fit models to match the noise profiles in actual human ratings (\autoref{sec:variance}).

Our second contribution is a reinforcement learning solution to bandit structured prediction and a study of its robustness to these simulated human ratings (\autoref{sec:method}).\footnote{Our code is at \url{https://github.com/khanhptnk/bandit-nmt} (in PyTorch).}
We combine an encoder-decoder architecture of machine translation~\cite{luong2015effective} with
the advantage actor-critic algorithm~\cite{mnih2016asynchronous}, yielding an approach that is
simple to implement but works on low-resource bandit machine translation.
Even with substantially restricted granularity, with high variance feedback, or with skewed rewards, this combination improves pre-trained models (\autoref{sec:results}).
In particular, under realistic settings of our noise parameters, the
algorithm's online reward and final held-out accuracies do not significantly degrade from a noise-free setting.

\section{Bandit Machine Translation} \label{sec:problem}

The bandit structured prediction problem \cite{daume15lols,sokolov2015coactive} is an extension of the contextual bandits problem \cite{kakade2008efficient,langford2008epoch} to structured prediction.
Bandit structured prediction operates over time $i=1 \dots K$ as:
\begin{enumerate}[nolistsep,noitemsep]
  \item World reveals context $\vec x^{(i)}$
  \item Algorithm predicts structured output $\hat{\vec y}^{(i)}$
  \item World reveals reward $R \left(\hat{\vec y}^{(i)}, \vec x^{(i)} \right)$
\end{enumerate}

We consider the problem of \emph{learning to translate from human ratings} in
a bandit structured prediction framework.
In each round, a translation model receives a source sentence $\vec x^{(i)}$, produces a
translation $\hat{\vec y}^{(i)}$, and receives a rating $R\left( \hat{\vec y}^{(i)}, \vec x^{(i)} \right)$
 from a human that reflects the quality of the translation.
We seek an algorithm that achieves high reward over $K$ rounds (high cumulative reward).
The challenge is that even though the model knows how good the translation is, 
it knows neither \emph{where} its mistakes are nor \emph{what} the ``correct'' translation looks like.
It must balance exploration (finding new good predictions) with exploitation (producing predictions it already knows are good).
This is especially difficult in a task like machine translation, where, for a twenty token sentence with a vocabulary size of $50k$, there are approximately $10^{94}$ possible outputs
, of which the algorithm gets to test exactly one.

\begin{figure}[t]
  \centering
  \includegraphics[width=\linewidth]{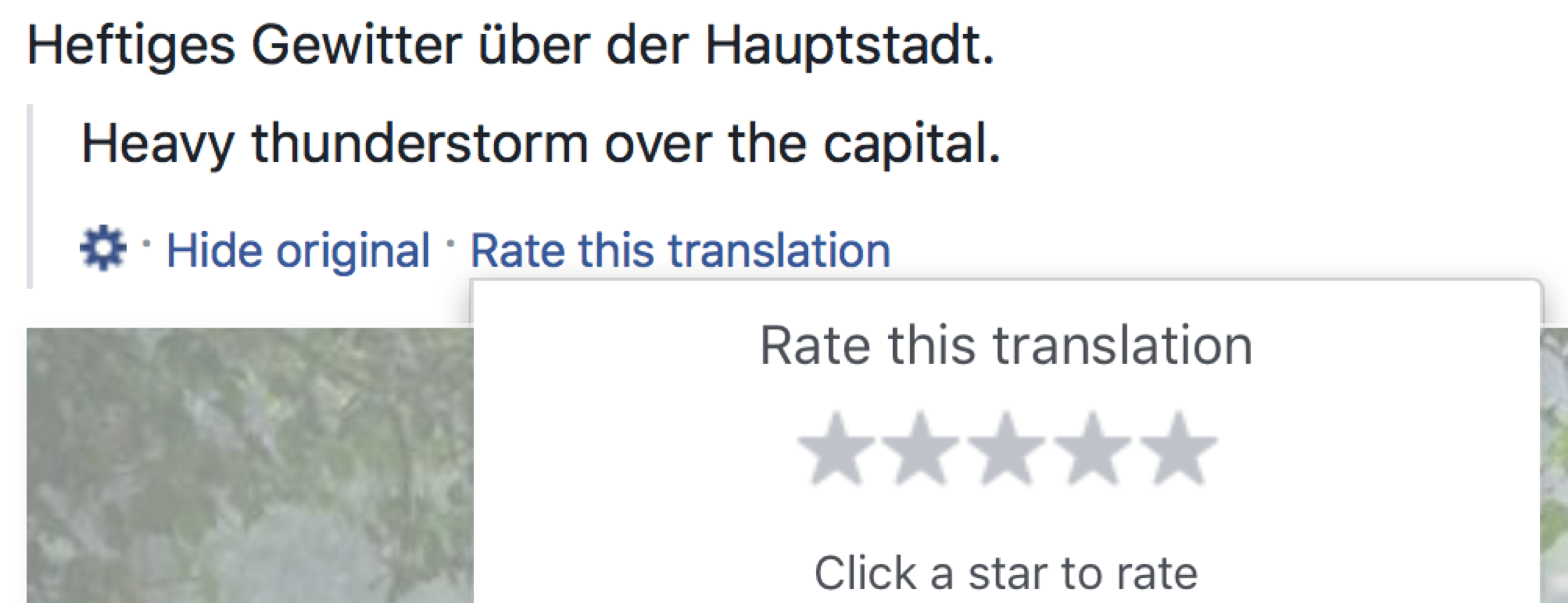}
  \caption{A translation rating interface provided by Facebook. Users see a sentence followed by its machined-generated translation and can give ratings from one to five stars. }
  \label{fig:facebook}
\end{figure}

Despite these challenges, learning from non-expert ratings is desirable.
In real-world scenarios, non-expert ratings are easy to collect but
other stronger forms of feedback are prohibitively expensive.
Platforms that offer translations can get quick feedback ``for free''
from their users to improve their systems (Figure \ref{fig:facebook}).
Even in a setting in which annotators are paid, it is much less expensive to
ask a bilingual speaker to provide a rating of a proposed translation
than it is to pay a professional translator to produce one from scratch.

\section{Effective Algorithm for Bandit MT} \label{sec:method}
This section describes the
neural machine translation architecture of our system (\autoref{sec:neural_mt}).
We formulate bandit neural machine translation as a reinforcement learning
problem (\autoref{sec:formulation}) and discuss why standard
actor-critic algorithms struggle with this problem
(\autoref{sec:why_ac_fail}).
Finally, we describe a more effective training approach based on the advantage
actor-critic algorithm (\autoref{sec:a2c}).

\subsection{Neural machine translation}
\label{sec:neural_mt}

Our neural machine translation (NMT) model is a neural encoder-decoder
that directly computes
the probability of translating a
target sentence $\vec y = (y_1, \cdots, y_m)$ from source sentence $\vec x$:
\begin{align}
    P_{\vec \theta}(\vec y \mid \vec x) = \prod_{t = 1}^m P_{\vec \theta}(y_t \mid \vec y_{<t}, \vec x)
  \end{align}
where $P_{\vec \theta}(y_t \mid \vec y_{<t}, \vec x)$ is the probability of
outputting the next word $y_t$ at time step $t$ given a translation prefix $\vec y_{<t}$ and
a source sentence $\vec x$.

We use an encoder-decoder NMT architecture with global attention~\cite{luong2015effective}, where
both the encoder and decoder are recurrent neural networks (RNN)
(see Appendix~A for a more detailed description).
These models are normally trained by supervised learning, but as reference translations are
not available in our setting, we use reinforcement learning methods, which only require
numerical feedback to function.

\subsection{Bandit NMT as Reinforcement Learning} \label{sec:formulation}

NMT generating process can be viewed as a Markov decision process
on a continuous state space.
The states are the hidden vectors $\vec h_t^{dec}$ generated by the
decoder.
The action space is the target language's vocabulary.

To generate a translation from a source sentence $\vec x$, an NMT model starts at an
initial state
$\vec h_0^{dec}$: a representation of $\vec x$ computed by the encoder.
At time step $t$, the model decides the next action to take
by defining a stochastic policy $P_{\vec \theta}(y_t \mid \vec y_{<t}, \vec x)$,
which is directly parametrized by the parameters $\vec \theta$ of the model.
This policy takes the current state $\vec h_{t - 1}^{dec}$ as input and produces a
probability distribution over all actions (target vocabulary words).
The next action $\hat{y}_t$ is chosen by taking $\arg\max$ or sampling from this distribution.
The model computes the next state $\vec h_t^{dec}$ by updating
the current state $\vec h_{t - 1}^{dec}$ by the action taken $\hat{y}_t$.


The objective of bandit NMT is to find a policy that maximizes the expected
reward of translations sampled from the model's policy:
\begin{equation}
  \max_{\vec \theta}\mathcal{L}_{pg}(\vec \theta) =
  \max_{\vec \theta} \mathbb{E}_{\substack{\vec x \sim D_{\textrm{tr}}\\\hat{\vec y} \sim P_{\vec \theta}(\cdot \mid \vec x)}}
  \Big[ R(\hat{\vec y}, \vec x)  \Big]
  \label{eqn:reward_max}
\end{equation} where $D_{tr}$ is the training set and $R$ is the reward function (the rater).\footnote{Our raters are \emph{stochastic}, but for simplicity we denote the reward as a function; it should be expected reward.}
We optimize this objective function with policy gradient methods.
For a fixed $\vec x$, the gradient of the objective in \autoref{eqn:reward_max} is:
\begin{align}
\label{eqn:pg_grad}
&  \nabla_{\vec \theta} \mathcal{L}_{pg}(\vec \theta) =
\mathbb{E}_{\hat{\vec y} \sim P_{\vec \theta}(\cdot)}
  \left[ R(\hat{\vec y}) \nabla_{\vec \theta} \log P_{\vec \theta}(\hat{\vec y})  \right]  \\
  &= \mathbb{E}_{\substack{\hat{\vec y} \sim P_{\vec \theta}(\cdot)}}
  \Big[
    \sum_{t = 1}^m \sum_{y_t}
    Q(\hat{\vec y}_{<t}, \hat{y}_t)
    \nabla_{\vec \theta}
    P_{\vec \theta}(\hat{y}_t \mid \hat{\vec y}_{<t})
    \Big] \nonumber
  \end{align} where $Q(\hat{\vec y}_{<t}, \hat{y}_t)$ is the expected future reward of 
$\hat{y}_t$ given the current prefix $\hat{\vec y}_{<t}$, then continuing
sampling from $P_{\vec \theta}$ to complete the translation:
  \begin{align}
  Q(\hat{\vec y}_{<t}, \hat{y}_t) &=
  \mathbb{E}_{\hat{\vec y}' \sim P_{\vec \theta}(\cdot \mid \vec x)}
   \left[
     \tilde{R}(\hat{\vec y}', \vec x)
  \right] \\
  \textrm{with } \tilde{R}(\hat{\vec y}', \vec x) & \equiv
   R(\hat{\vec y}', \vec x)
   \mathbbm{1}
   \left\{ \hat{\vec y}'_{<t} = \hat{\vec y}_{<t}, \hat{y}'_t = \hat{y}_t \right\} \nonumber
\end{align}
$\mathbbm{1}\{\cdot\}$ is the indicator function, which returns 1 if the logic inside the bracket is true and returns 0 otherwise.

The gradient in \autoref{eqn:pg_grad} requires rating all possible
translations, which is not feasible in bandit NMT.
Na\"ive Monte Carlo reinforcement learning methods such as REINFORCE~\cite{williams1992simple}
estimates $Q$ values by sample means but yields very high variance when the action space is large,
leading to training instability.

\subsection{Why are actor-critic algorithms not effective for bandit NMT?}
\label{sec:why_ac_fail}

Reinforcement learning methods that rely on function approximation are preferred when tackling
bandit structured prediction with a large action space because they can capture
similarities between structures and generalize to unseen regions of the structure space.
The actor-critic algorithm~\cite{konda1999actor} uses function approximation to directly model
the $Q$ function, called the \emph{critic} model. 
In our early attempts on bandit NMT, we adapted the actor-critic algorithm for NMT 
in \citet{bahdanau2016actor}, which employs the algorithm in a supervised learning setting.
Specifically, while an encoder-decoder critic model $Q_{\vec \omega}$
as a substitute for the
true $Q$ function in \autoref{eqn:pg_grad} enables
taking the full sum inside the expectation (because
the critic model can be queried with any state-action pair), we are unable to obtain reasonable results with this approach.


Nevertheless, insights into why this approach fails on our problem 
explains the effectiveness of the approach discussed in the next section.
There are two properties in \citet{bahdanau2016actor} that our problem lacks but are key
elements for a successful actor-critic.
The first is access to reference translations: while the critic model is able
to observe reference translations during training in their setting,
bandit NMT assumes those are never available.
The second is per-step rewards: while the reward function in their setting
is known and can be exploited to compute immediate rewards after taking each action,
in bandit NMT, the actor-critic algorithm struggles with credit assignment
because it only receives reward when a translation is completed.
\citet{bahdanau2016actor} report that
the algorithm degrades if rewards are delayed until the end,
consistent with our observations.

With an enormous action space of bandit NMT, approximating gradients with the $Q$ critic model 
induces biases and potentially drives the model to wrong optima.
Values of rarely taken actions are often overestimated without an explicit constraint between $Q$ values of actions (e.g., a sum-to-one constraint).
\citet{bahdanau2016actor} add an ad-hoc regularization term to the loss function to mitigate this issue and further stablizes the algorithm with a delay update scheme, but at the same time introduces extra tuning hyper-parameters.


\subsection{Advantage Actor-Critic for Bandit NMT} \label{sec:a2c}

We follow the approach of advantage actor-critic~\cite[A2C]{mnih2016asynchronous} and combine it with the neural encoder-decoder architecture.
The resulting algorithm---which we call NED-A2C---approximates the gradient in \autoref{eqn:pg_grad}
by a single sample $\hat{\vec y} \sim P(\cdot \mid \hat{\vec x})$ and centers the
reward $R(\hat{\vec y})$ using
the state-specific expected future reward $V(\hat{\vec y}_{<t})$ to reduce variance:
\begin{align}
\label{eqn:a2c_grad}
  \nabla_{\vec \theta} \mathcal{L}_{pg}(\vec \theta) & \approx
  \sum_{t = 1}^m
  \bar{R}_t(\hat{\vec y})
  \nabla_{\vec \theta}
  \log P_{\vec \theta}(\hat{y}_t \mid \hat{\vec y}_{<t}) \\
  \textrm{with }\bar{R}_t(\hat{\vec y}) &\equiv R(\hat{\vec y}) - V(\hat{\vec y}_{<t}) \nonumber \\
  V(\hat{\vec y}_{<t}) &\equiv \mathbb{E}_{\hat{y}'_t \sim P(\cdot \mid \hat{\vec y}_{<t})}
  \left[  Q(\hat{\vec y}_{<t}, \hat{y}'_t) \right] \nonumber
\end{align}

We train a separate attention-based encoder-decoder model $V_{\vec \omega}$ to estimate
$V$ values.
This model encodes a source sentence $\vec x$ and decodes a sampled
translation $\hat{\vec y}$.
At time step $t$, it computes
$V_{\vec \omega}(\hat{\vec y}_{<t}, \vec x) = \vec w^{\top} \vec h_t^{crt}$,
where $\vec h^{crt}_t$ is the current decoder's hidden vector and
$\vec w$ is a learned weight vector.
The critic model minimizes the MSE between its estimates and the true
values:
\begin{align}
 \label{eqn:vnet}
  \mathcal{L}_{crt}(\vec \omega) & = \mathbb{E}_{\substack{\vec x \sim D_{\textrm{tr}}\\\hat{\vec y} \sim P_{\vec \theta}(\cdot \mid \vec x)}} \left[
  \sum_{t = 1}^m L_t(\hat{\vec y}, \vec x) \right] \\
  \textrm{with } L_t(\hat{\vec y}, \vec x) & \equiv 
  \left[ 
   V_{\vec \omega}(\hat{\vec y}_{<t}, \vec x) - R(\hat{\vec y}, \vec x)   \right]^2. \nonumber
\end{align} We use a gradient approximation to update $\vec \omega$ for a fixed $\vec x$ and $\hat{\vec y} \sim P(\cdot \mid \hat{\vec x})$:
\begin{equation}
  \nabla_{\vec \omega} \mathcal{L}_{crt}(\vec \omega) \approx
 \sum_{t = 1}^m \left[ 
  V_{\vec \omega}(\hat{\vec y}_{<t}) - R(\hat{\vec y}) 
\right]
                      \nabla_{\vec \omega} V_{\vec \omega}(\hat{\vec y}_{<t})
 \label{eqn:critic_grad}
\end{equation} 

NED-A2C is better suited for problems with a large action space and has other
advantages over actor-critic.
For large action spaces, approximating gradients using the $V$ critic model induces 
lower biases than using the $Q$ critic model.
As implied by its definition, the $V$ model is robust to biases incurred by
rarely taken actions since rewards of those actions are weighted by very small
probabilities in the expectation.
In addition, the $V$ model has a much smaller
number of parameters and thus is more sample-efficient and more stable to train than the $Q$ model.
These attractive properties were not studied in A2C's original
paper~\cite{mnih2016asynchronous}.

\begin{algorithm}
  \caption{The NED-A2C algorithm for bandit NMT.}
  \label{alg:a2c}
\begin{algorithmic}[1]
  \small
  \FOR{$i = 1 \cdots K$}
  \STATE receive a source sentence $\vec x^{(i)}$
  \STATE sample a translation: $\hat{\vec y}^{(i)} \sim P_{\vec \theta}(\cdot \mid \vec x^{(i)})$
  \STATE receive reward $R(\hat{\vec y}^{(i)}, \vec x^{(i)})$
  \STATE update the NMT model using \autoref{eqn:a2c_grad}.
  \STATE update the critic model using \autoref{eqn:critic_grad}.
  \ENDFOR
\end{algorithmic}
\end{algorithm}

Algorithm \ref{alg:a2c} summarizes NED-A2C for bandit NMT.
For each $\vec x$, we draw a single sample $\hat{\vec y}$ from the NMT model, which
is used for both estimating gradients of the NMT model and the critic model.
We run this algorithm with mini-batches of $\vec x$ and aggregate
gradients over all $\vec x$ in a mini-batch for each update.
Although our focus is on bandit NMT, this algorithm naturally works with any
bandit structured prediction problem.


\section{Modeling Imperfect Ratings} \label{sec:noise_model}
\begin{figure}[t]
  \centering
  \includegraphics[width=0.8\linewidth]{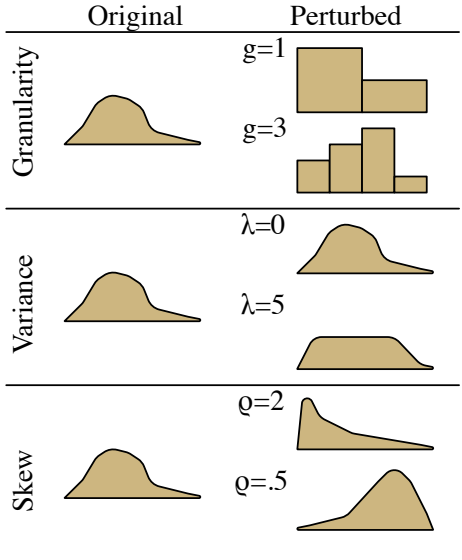}
  \caption{Examples of how our perturbation functions change the ``true''
    feedback distribution (left) to ones that better capture features
    found in human feedback (right).}
  \label{fig:perturbation}

\end{figure}

Our goal is to establish the feasibility of using \emph{real} human feedback to optimize a machine translation system,
in a setting where one can collect \emph{expert} feedback as well as a setting in which one only collects \emph{non-expert} feedback.
In all cases, we consider the expert feedback to be the ``gold standard'' that we wish to optimize.
To establish the feasibility of driving learning from human feedback \emph{without} doing a full, costly user study, we begin with a simulation study.
The key aspects (\autoref{fig:perturbation}) of human feedback we capture are:
(a) mismatch between training objective and feedback-maximizing objective,
(b) human ratings typically are binned (\autoref{sec:granular}),
(c) individual human ratings have high variance (\autoref{sec:variance}),
and (d) non-expert ratings can be skewed with respect to expert ratings (\autoref{sec:skew}).

In our simulated study, we begin by modeling gold standard human ratings using add-one-smoothed sentence-level \bleu~\cite{chen2014systematic}.\footnote{``Smoothing 2'' in~\citet{chen2014systematic}. We also add one to lengths when computing the brevity penalty.}
Our evaluation criteria, therefore, is average sentence-\bleu over the run of our algorithm.
However, in any realistic scenario, human feedback will vary from its average, and so the reward that our algorithm receives will be a \emph{perturbed} variant of sentence-\bleu.
In particular, if the sentence-\bleu score is $s \in [0, 1]$, the algorithm will only observe $s' \sim \textrm{pert}(s)$, where $\textrm{pert}$ is a perturbation distribution.
Because our reference machine translation system is pre-trained using log-likelihood, there is already an (a) mismatch between training objective and feedback, so we focus on (b-d) below.

\subsection{Humans Provide Granular Feedback} \label{sec:granular}

When collecting human feedback, it is often more effective to collect discrete \emph{binned} scores.
A classic example is the Likert scale for human agreement \cite{likert:1932} or star ratings for product reviews.
Insisting that human judges provide continuous values (or feedback at too fine a
granularity) can demotivate raters without improving rating quality \cite{Preston20001}.

To model granular feedback, we use a simple rounding procedure.
Given an integer parameter $g$ for degree of granularity, we define: 
%
\begin{align}
  \textrm{pert}^{\textrm{gran}}(s;g)
  &= \frac 1 g \textrm{round}(g s)
\end{align}
This perturbation function divides the range of possible outputs into $g+1$ bins.
For example, for $g=5$, we obtain bins $[0,0.1)$, $[0.1, 0.3)$, $[0.3, 0.5)$, $[0.5, 0.7)$, $[0.7, 0.9)$ and $[0.9,1.0]$. Since most sentence-\bleu scores are much closer to zero than to one, many of the larger bins are frequently vacant.

\subsection{Experts Have High Variance} \label{sec:variance}

Human feedback has high variance around its expected value.
A natural goal for a variance model of human annotators is to simulate---as closely as possible---how human raters actually perform.
We use human evaluation data recently collected as part of the WMT shared task~\cite{graham2017can}.
The data consist of 7200 sentences multiply annotated by giving non-expert annotators on Amazon Mechanical Turk a reference sentence and a \emph{single} system translation, and asking the raters to judge the adequacy of the translation.\footnote{Typical machine translation evaluations evaluate pairs and ask annotators to choose which is better.} 

\begin{figure}[t]
  \centering
  \includegraphics[width=\linewidth,clip=true,trim=35 22 61 311]{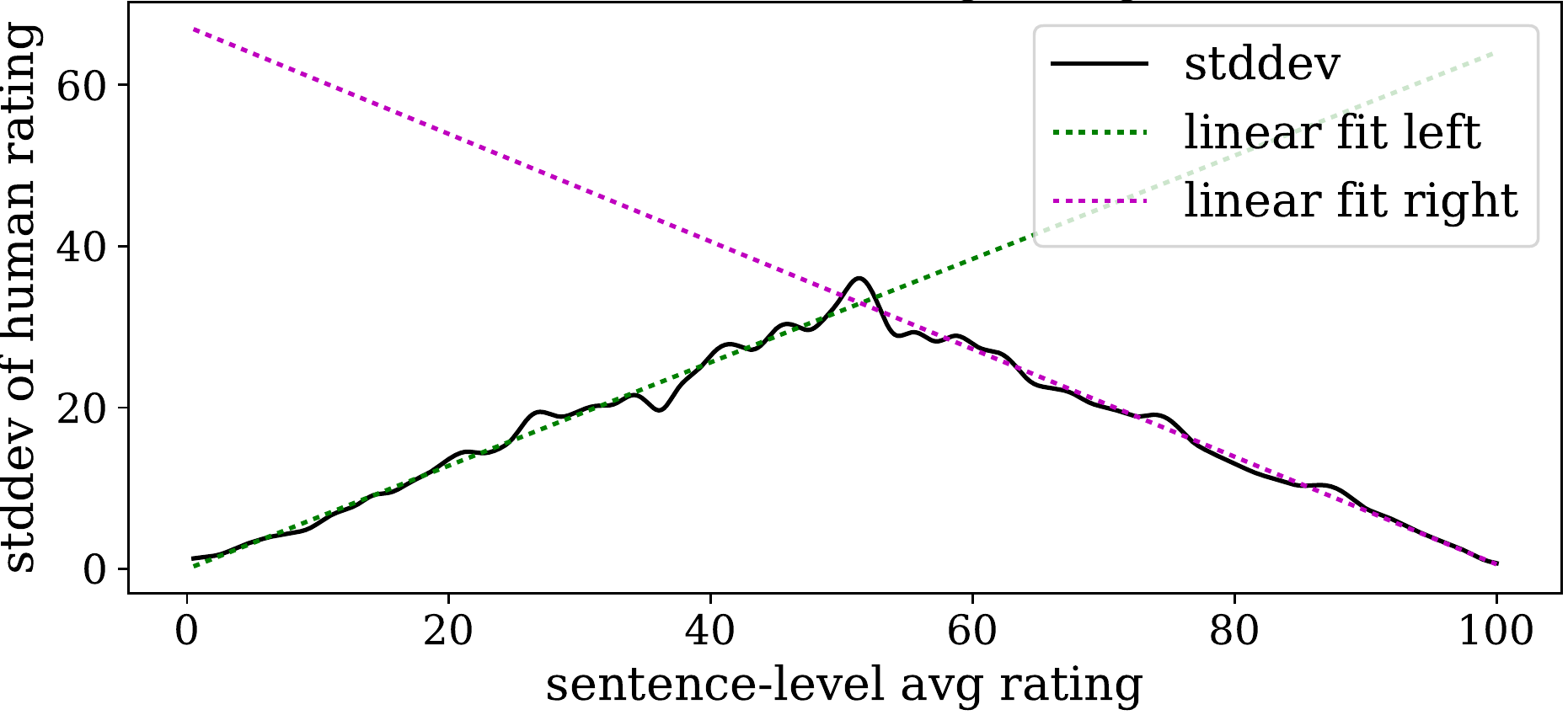}
  \caption{Average rating (x-axis) versus a kernel density estimate of the variance of human ratings around that mean, with linear fits. 
  Human scores vary more around middling judgments than extreme judgments.}
  \label{fig:yvettedata}
\end{figure}

From these data, we treat the \emph{average} human rating as the ground truth and consider how individual human ratings vary around that mean.
To visualize these results with kernel density estimates (standard normal kernels) of the \emph{standard deviation}.
\autoref{fig:yvettedata} shows the mean rating (x-axis) and the deviation of the human ratings (y-axis) at each mean.\footnote{A current limitation of this model is that the simulated noise is i.i.d. conditioned on the rating (homoscedastic noise). While this is a stronger and more realistic model than assuming no noise, real noise is likely heteroscedastic: dependent on the input.}
As expected, the standard deviation is small at the extremes and large in the middle (this is a bounded interval), with a fairly large range in the middle:
a translation whose average score is $50$ can get human evaluation scores anywhere between $20$ and $80$ with high probability.
We use a linear approximation to define our variance-based perturbation function as a Gaussian distribution, which is parameterized by a scale $\lambda$ that grows or shrinks the variances (when $\lambda=1$ this exactly matches the variance in the plot).
\renewcommand{\brack}[1]{\left\{\begin{array}{ll}#1\end{array}\right.}
\begin{align}
  \textrm{pert}^{\textrm{var}}(s;\lambda)
  &= \textrm{Nor}\left(s, \lambda \sigma(s)^2\right) \\
   \sigma(s) &= \brack{~~~0.64 s - 0.02 & \textrm{if } s < 50 \\
                         -0.67 s + 67.0 & \textrm{otherwise} } \nonumber
\end{align}


\subsection{Non-Experts are Skewed from Experts} \label{sec:skew}

The preceding two noise models assume that the reward closely models the value we want to optimize (has the same mean).
This may not be the case with non-expert ratings.
Non-expert raters are skewed both for reinforcement learning \cite{thomaz2006reinforcement,thomaz2008teachable,loftin2014strategy} and recommender systems \cite{herlocker2000explaining,adomavicius2012impact}, but are typically bimodal: some are harsh (typically provide very low scores, even for ``okay'' outputs) and some are motivational (providing high scores for ``okay'' outputs).

We can model both harsh and motivations raters with a simple deterministic skew perturbation function, parametrized by a scalar $\rho \in [0,\infty)$:
\begin{align}
  \textrm{pert}^{\textrm{skew}}(s;\rho)
  &= s^\rho
\end{align}
For $\rho > 1$, the rater is harsh; for $\rho < 1$, the rater is motivational.




\section{Experimental Setup} \label{sec:experiment}
\begin{table}[t]
  \centering
  \small
\label{my-label}
\begin{tabular}{lll}
\toprule
                    & De-En & Zh-En \\ \midrule
Supervised training & 186K  & 190K  \\ 
Bandit training     & 167K  & 165K  \\ 
Development         & 7.7K  & 7.9K  \\ 
Test                & 9.1K  & 7.4K  \\ 
\bottomrule
\end{tabular}
\caption{Sentence counts in data sets.}
\label{tab:data}
\end{table}

We choose two language pairs from different language families with different typological properties: German-to-English and (De-En) and Chinese-to-English (Zh-En).
We use parallel transcriptions of TED talks for these pairs of languages 
from the machine translation track of the IWSLT 2014 and 2015~\cite{cettolo2014report,cettolo2015iwslt,cettoloEtAl:EAMT2012}.
For each language pair, we split its data into four sets for supervised training, 
bandit training, development and testing (Table \ref{tab:data}).
For English and German, we tokenize and clean sentences using Moses~\cite{koehn2007moses}.
For Chinese, we use the Stanford Chinese word segmenter \cite{chang08chinese} to segment sentences and tokenize.
We remove all sentences with length greater than 50, resulting in an average 
sentence length of 18.
We use IWSLT 2015 data for supervised training and development, IWSLT 2014 data 
for bandit training and previous years' development and evaluation data for testing.\footnote{Over 95\% of the bandit learning set's sentences are seen during supervised learning. Performance gain on this set mainly reflects how well a model leverages weak learning signals (ratings) to improve previously made predictions. Generalizability is measured by performance gain on the test sets, which do not overlap the training sets.}

\subsection{Evaluation Framework}

For each task, we first use the supervised training set to pre-train 
a reference NMT model using supervised learning.
On the same training set, we also pre-train the critic model with 
translations sampled from the pre-trained NMT model. 
Next, we enter a bandit learning mode where our models only observe the source
sentences of the bandit training set. 
Unless specified differently, we train the NMT models with NED-A2C for one pass 
over the bandit training set.
If a perturbation function is applied to \persentence scores, it is only applied
in this stage, not in the pre-training stage. 

We measure the \emph{improvement} $\Delta S$ of
an evaluation metric $S$ due to bandit training: $\Delta S = S_{A2C} - S_{ref}$, where $S_{ref}$ is the metric computed on the reference models
and $S_{A2C}$ is the metric computed on models trained with NED-A2C.
Our primary interest is \emph{\persentence}: average sentence-level \bleu of translations that
    are sampled and scored during the bandit learning pass. 
    This metric represents average expert ratings, which we want to optimize for 
    in real-world scenarios. 
We also measure \emph{\heldout}: corpus-level \bleu on an unseen test set, where
    translations are greedily decoded by the NMT models. 
    This shows how much our method improves translation 
    quality, since corpus-level \bleu correlates better with human judgments
    than sentence-level \bleu. 

Because of randomness due to both the random sampling in the model for ``exploration'' as well as the randomness in the reward function, we repeat each experiment five times and 
report the mean results with 95\% confidence intervals.

\subsection{Model configuration}

Both the NMT model and the critic model are encoder-decoder models with global 
attention~\cite{luong2015effective}.
The encoder and the decoder are unidirectional single-layer LSTMs. 
They have the same word embedding size and LSTM hidden size of 500.
The source and target vocabulary sizes are both 50K.
We do not use dropout in our experiments.
We train our models by the Adam optimizer \cite{kingma14adam} with 
$\beta_1 = 0.9, \beta_2 = 0.999$ and a batch size of 64.
For Adam's $\alpha$ hyperparameter, we use $10^{-3}$ during pre-training and $10^{-4}$ during
bandit learning (for both the NMT model and the critic model). 
During pre-training, starting from the fifth pass, we decay $\alpha$ by a factor of 0.5 when perplexity on the development set increases.
The NMT model reaches its highest corpus-level \bleu on the development set after 
ten passes through the supervised training data, while the critic model's training error stabilizes 
after five passes.
The training speed is 18s/batch for supervised pre-training and 41s/batch for
training with the NED-A2C algorithm. 


\section{Results and Analysis} \label{sec:results}
In this section, we describe the results of our experiments,
broken into the following questions:
how NED-A2C improves reference models (\autoref{sec:persentence});
the effect the three perturbation functions have on the algorithm 
(\autoref{sec:perturb_results}); and whether the algorithm improves a corpus-level
metric that corresponds well with human judgments (\autoref{sec:heldout_results}).

\subsection{Effectiveness of NED-A2C under Un-perturbed Bandit Feedback}
\label{sec:persentence}

\begin{table*}[!t]
  \small
\centering
\begin{tabular}{l|ccc|ccc}
\toprule
       & \multicolumn{3}{c|}{De-En}          & \multicolumn{3}{c}{Zh-En}   \\
       & Reference  & $\Delta_{sup}$ & $\Delta_{A2C}$     
       & Reference  & $\Delta_{sup}$ & $\Delta_{A2C}$     \\ \midrule
       \multicolumn{7}{c}{Fully pre-trained reference model} \\ \midrule
\persentence & 38.26 $\pm$ 0.02 & 0.07 $\pm$ 0.05 & \textbf{2.82 $\pm$ 0.03} 
             & 32.79 $\pm$ 0.01 & 0.36 $\pm$ 0.05 & \textbf{1.08 $\pm$ 0.03} \\ 
\heldout & 24.94 $\pm$ 0.00 & 1.48 $\pm$ 0.00 & \textbf{1.82 $\pm$ 0.08} 
         & 13.73 $\pm$ 0.00 & \textbf{1.18 $\pm$ 0.00} & 0.86 $\pm$ 0.11 \\ \midrule
       \multicolumn{7}{c}{Weakly pre-trained reference model} \\ \midrule
\persentence & 19.15 $\pm$ 0.01 & 2.94 $\pm$ 0.02 & \textbf{7.07 $\pm$ 0.06} 
             & 14.77 $\pm$ 0.01 & 1.11 $\pm$ 0.02 & \textbf{3.60 $\pm$ 0.04} \\ 
  \heldout   & 19.63 $\pm$ 0.00 & \textbf{3.94 $\pm$ 0.00} & 1.61 $\pm$ 0.17 
             & 9.34 $\pm$ 0.00 & \textbf{2.31 $\pm$ 0.00}  & 0.92 $\pm$ 0.13 \\ \midrule
\end{tabular}
\caption{Translation scores and improvements based on a single round of un-perturbed bandit feedback. \persentence and \heldout are not comparable: the former is sentence-\bleu, the latter is corpus-\bleu.}
\label{tab:clean_bleu}
\end{table*}

We evaluate our method in an ideal setting where \emph{un-perturbed} \persentence simulates
ratings during both training and evaluation (Table \ref{tab:clean_bleu}).

\paragraph{Single round of feedback.}
In this setting, our models only observe each source sentence once and 
before producing its translation.
On both De-En and Zh-En, NED-A2C improves \persentence of
reference models 
after only a single pass (+2.82 and +1.08 respectively). 

\paragraph{Poor initialization.}
Policy gradient algorithms have difficulty improving from poor 
initializations, especially on problems with a large action space, because 
they use model-based exploration, which is ineffective when 
most actions have equal probabilities~\cite{bahdanau2016actor,ranzato2015sequence}.
To see whether NED-A2C has this problem, we repeat the experiment with the same
setup but with reference models pre-trained for only a single pass.
Surprisingly, NED-A2C is highly effective at improving these poorly trained models 
(+7.07 on De-En and +3.60 on Zh-En in \persentence).

\paragraph{Comparisons with supervised learning.}
To further demonstrate the effectiveness of NED-A2C, we compare it
with training the reference models with supervised learning for a single pass on the bandit training set.
Surprisingly, observing ground-truth translations barely improves the models in \persentence 
when they are fully trained (less than +0.4 on both tasks). 
A possible explanation is that the models have already reached full capacity and do not benefit 
from more examples.\footnote{This result may vary if the domains of the supervised learning set and the bandit training set are dissimilar. Our training data are all TED talks. }
NED-A2C further enhances the models because it eliminates the mismatch between the supervised training objective and the evaluation objective. 
On weakly trained reference models, NED-A2C also significantly outperforms supervised learning ($\Delta$\persentence of NED-A2C is over three times as large as those of supervised learning).

\begin{figure}[t]
  \centering
  \includegraphics[width=0.9\linewidth]{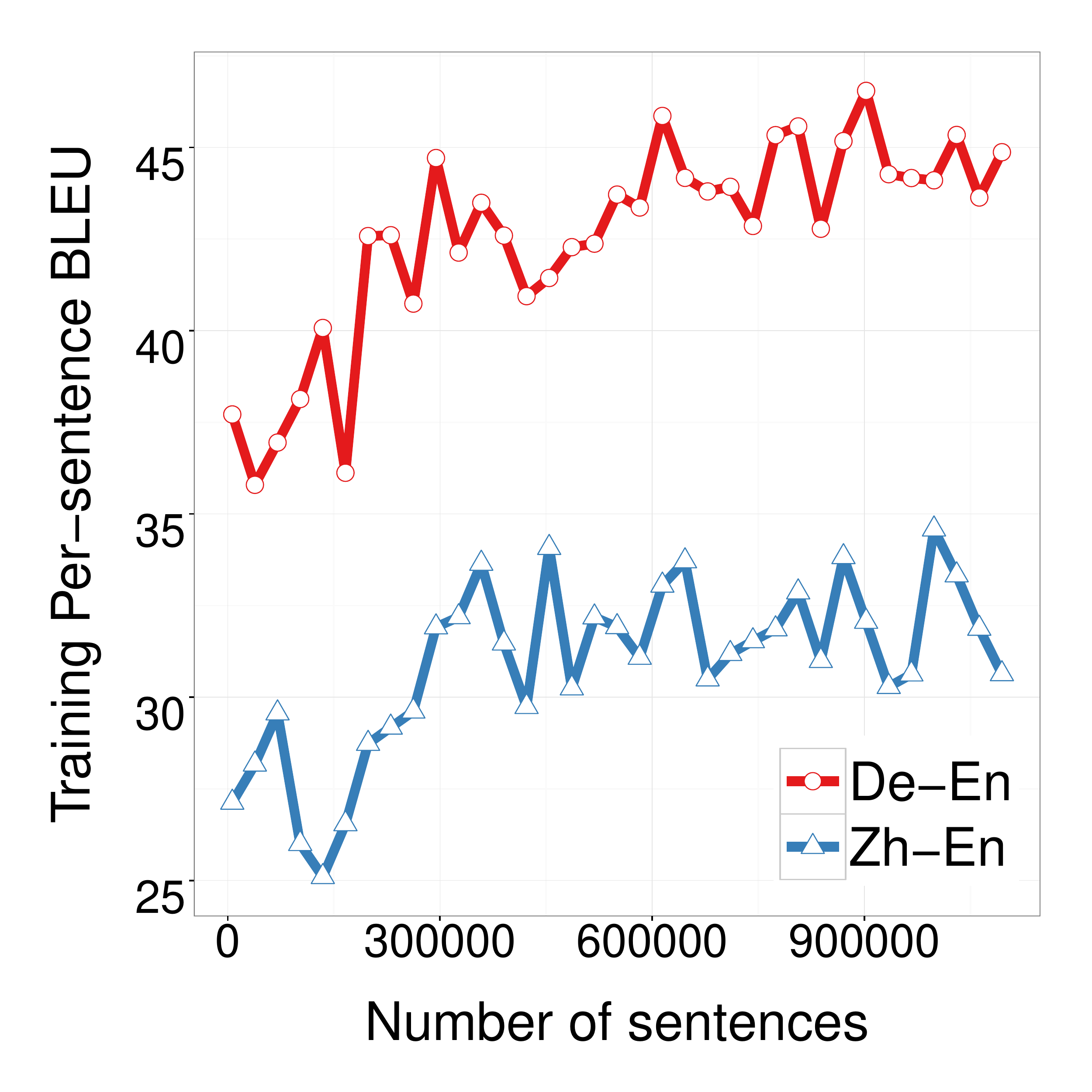}
  \caption{Learning curves of models trained with NED-A2C for five epochs.}
  \label{fig:curves}
\end{figure}

\paragraph{Multiple rounds of feedback.}
We examine if NED-A2C can improve the models even further with multiple rounds of feedback.\footnote{The ability to receive feedback on the same example multiple times might 
not fit all use cases though.} 
With supervised learning, the models can memorize the reference translations
but, in this case, the models have to be able to exploit and explore effectively. 
We train the models with NED-A2C for five passes and observe a much more significant 
$\Delta$\persentence than training for a single pass in both pairs of language 
(+6.73 on De-En and +4.56 on Zh-En) (\autoref{fig:curves}).

\begin{figure*}[!t]
  \centering
  \subcaptionbox{Granularity}[.3\linewidth][c]{%
    \includegraphics[width=.3\linewidth,clip=true,trim=28 15 28 8]{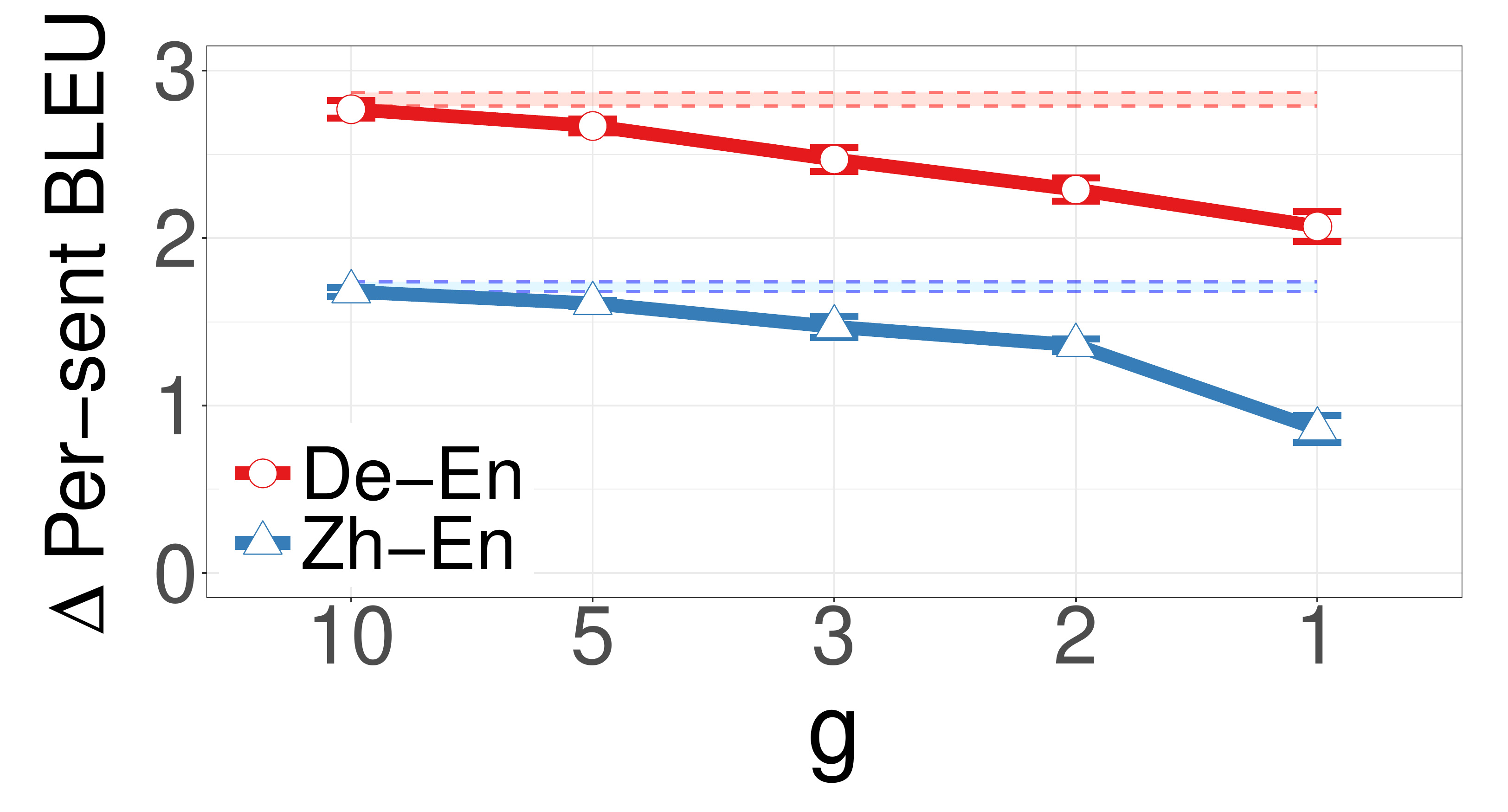}\vspace{-2mm}}
  \subcaptionbox{Variance}[.3\linewidth][c]{%
    \includegraphics[width=.3\linewidth,clip=true,trim=28 15 28 8]{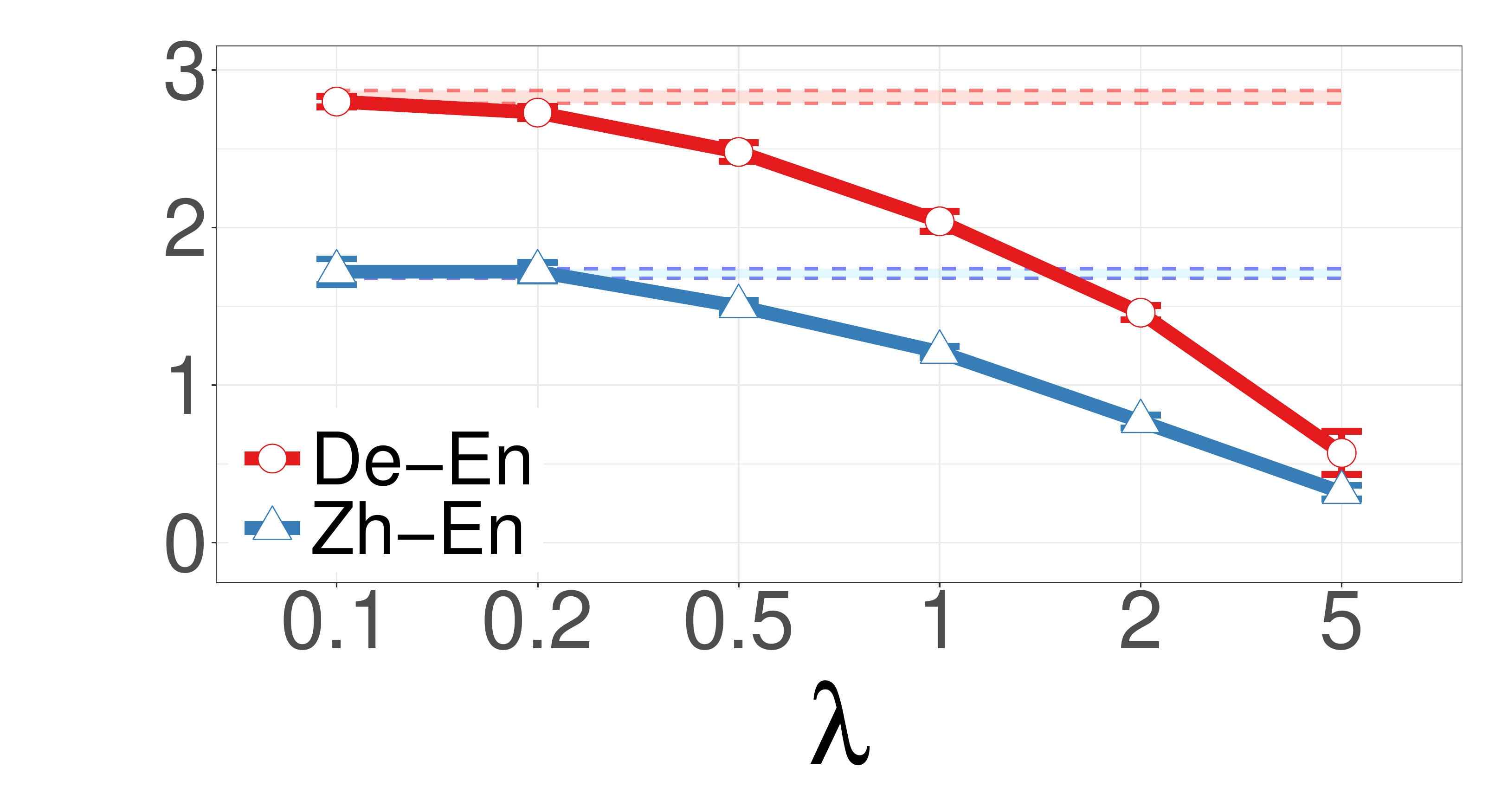}\vspace{-2mm}}
  \subcaptionbox{Skew}[.3\linewidth][c]{%
    \includegraphics[width=.3\linewidth,clip=true,trim=28 15 28 8]{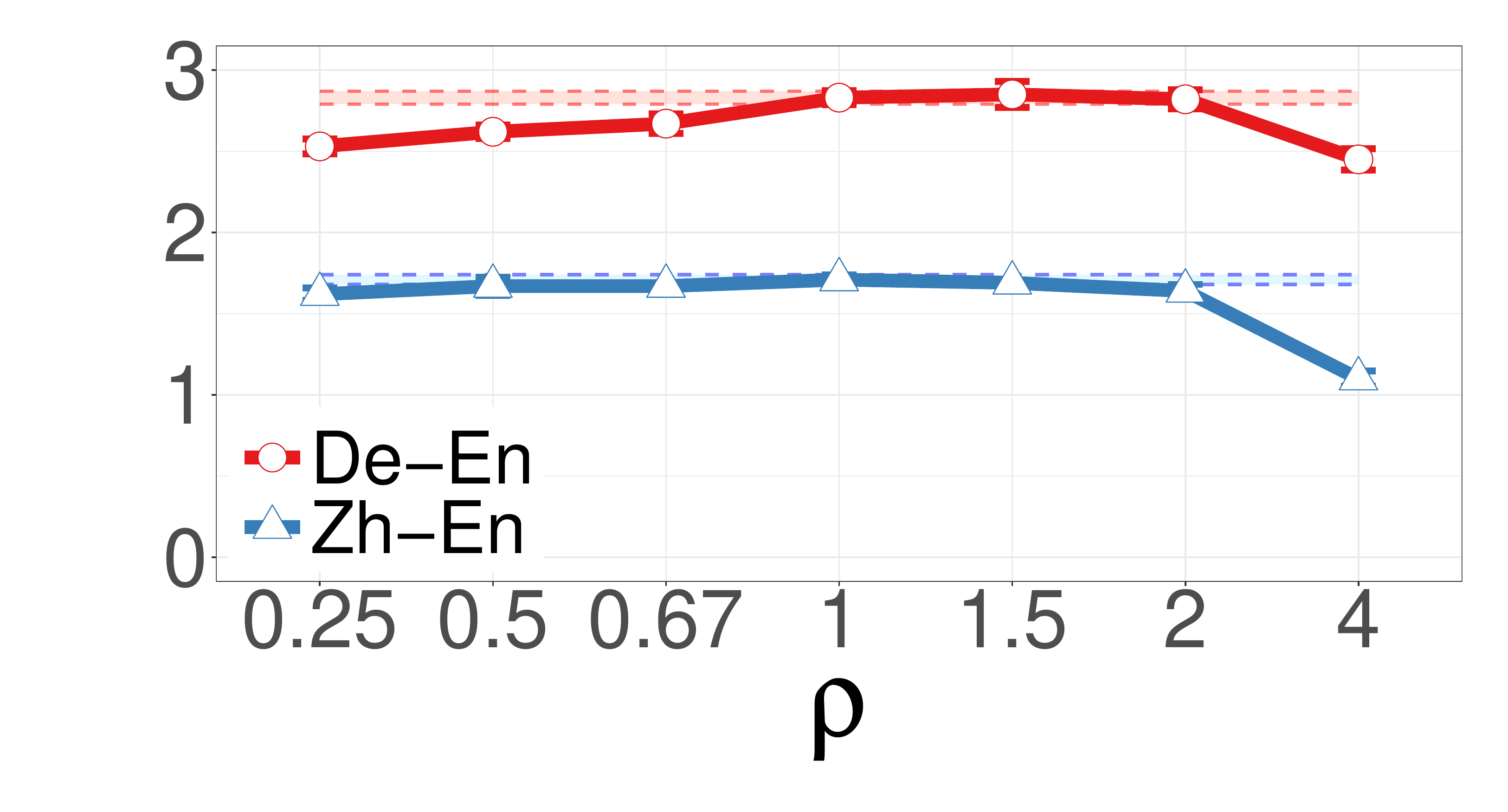}\vspace{-2mm}}
  

  \subcaptionbox{Granularity}[.3\linewidth][c]{%
    \includegraphics[width=.3\linewidth,clip=true,trim=28 15 28 8]{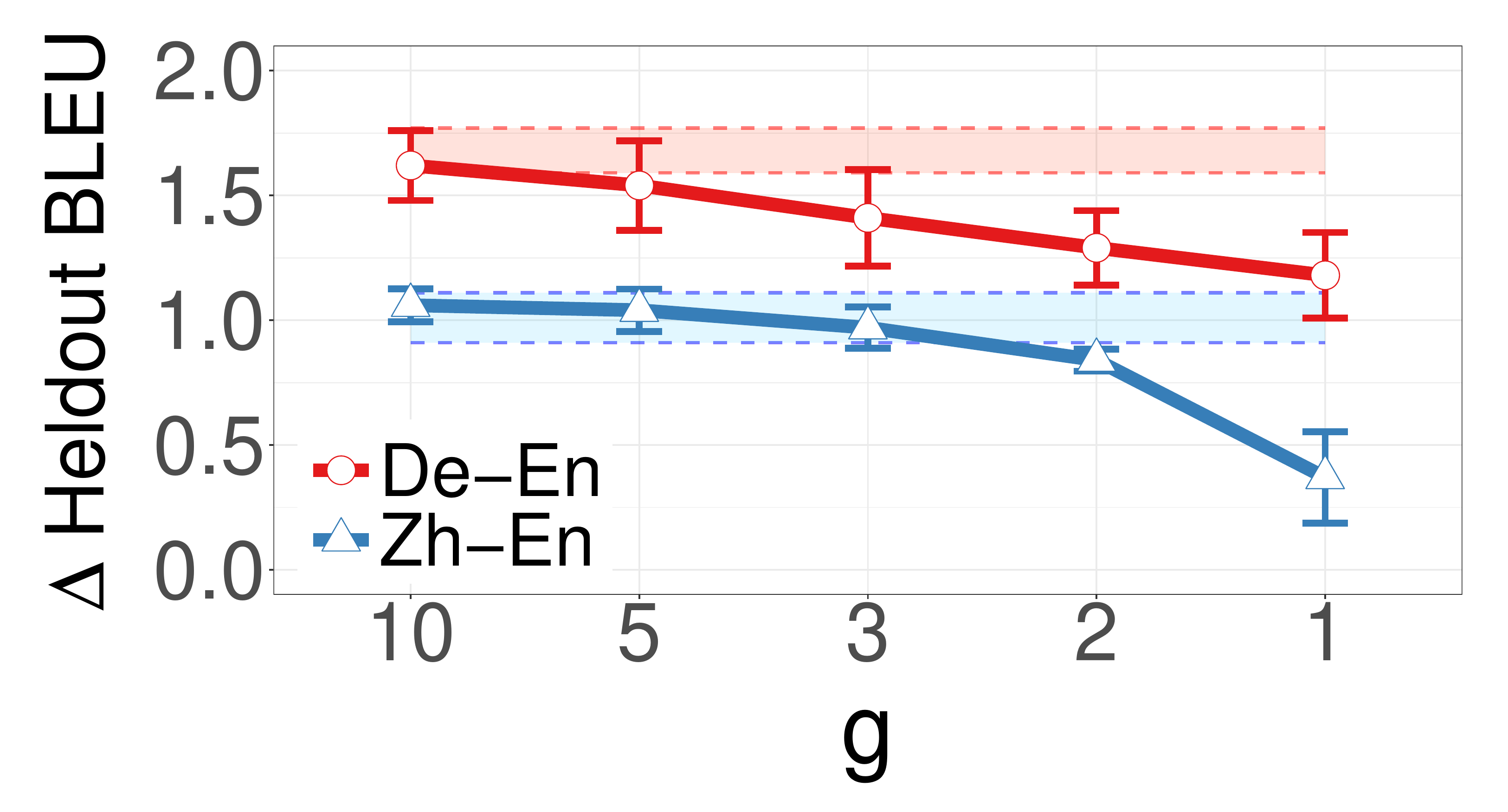}\vspace{-2mm}}
  \subcaptionbox{Variance}[.3\linewidth][c]{%
    \includegraphics[width=.3\linewidth,clip=true,trim=28 15 28 8]{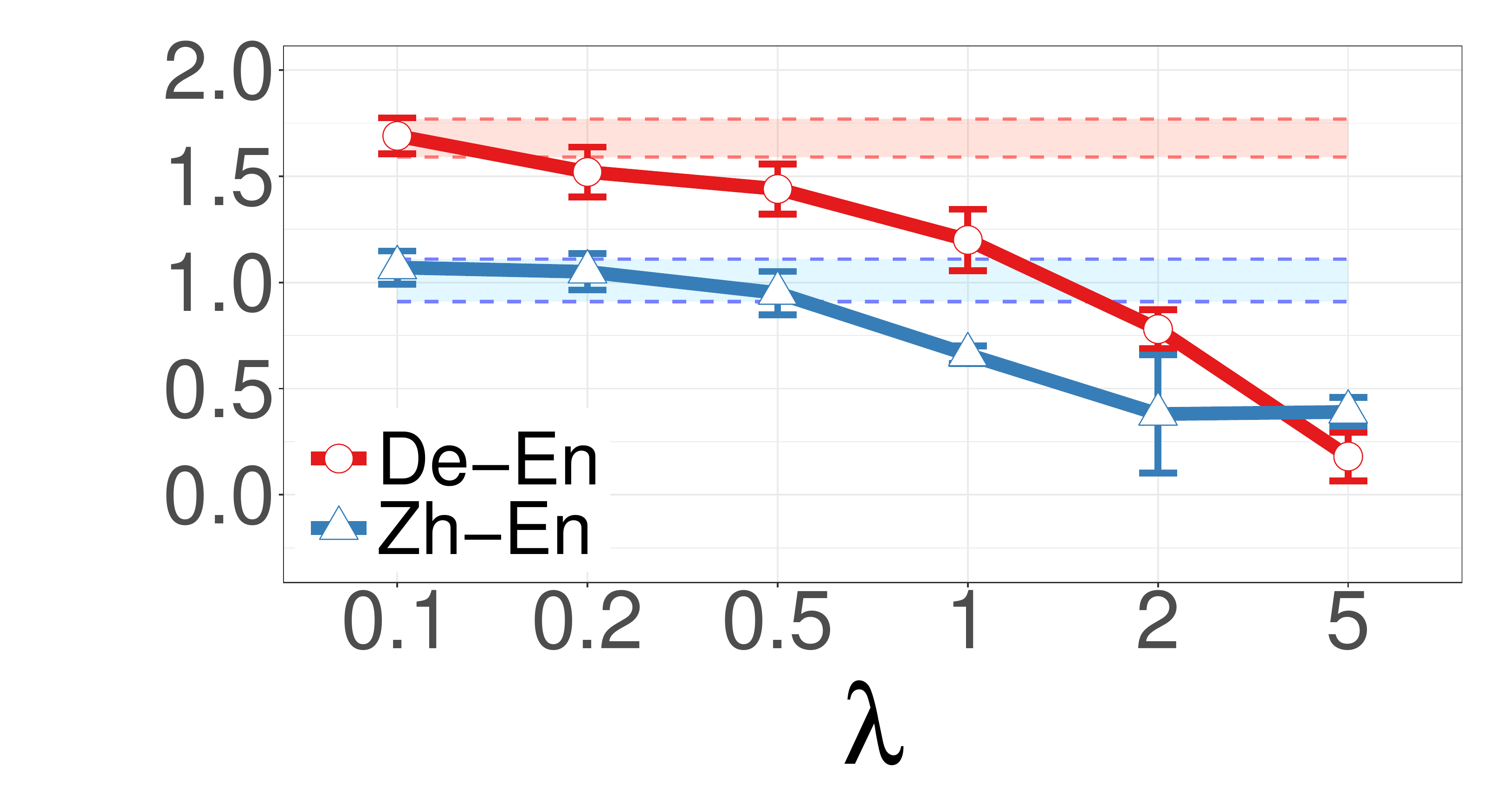}\vspace{-2mm}}
  \subcaptionbox{Skew}[.3\linewidth][c]{%
    \includegraphics[width=.3\linewidth,clip=true,trim=28 15 28 8]{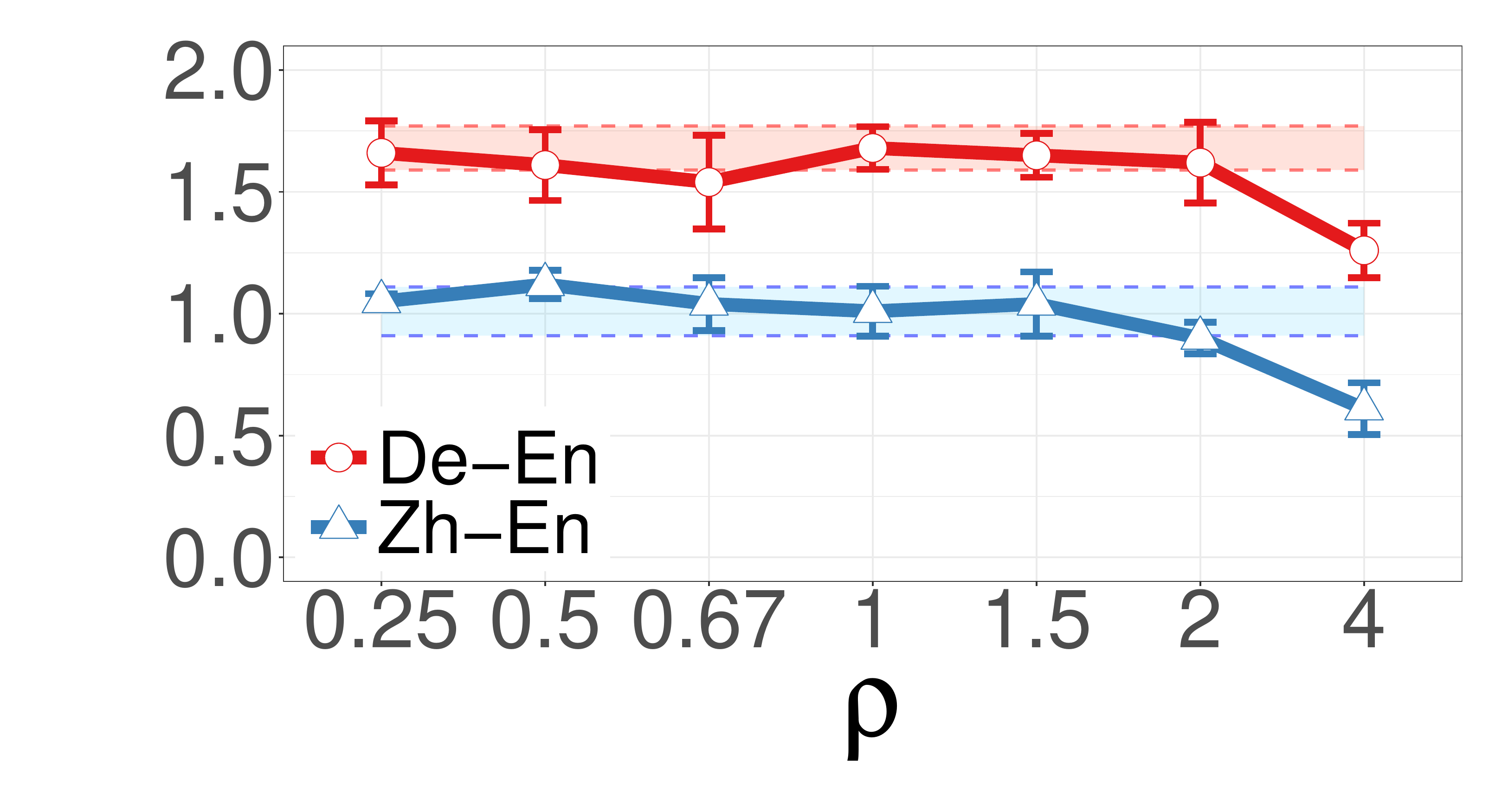}\vspace{-2mm}}
  \caption{Performance gains of NMT models trained with NED-A2C in \persentence 
    (top row) and in \heldout(bottom row) under various degrees of granularity, 
    variance, and skew of scores. 
    Performance gains of models trained with un-perturbed scores are within the
    shaded regions. 
  }
  \label{fig:noise}
\end{figure*}

\subsection{Effect of Perturbed Bandit Feedback}
\label{sec:perturb_results}
We apply perturbation functions defined in \autoref{sec:granular} to \persentence scores and 
use the perturbed scores as rewards during bandit training (\autoref{fig:noise}).

\paragraph{Granular Rewards.} We discretize raw \persentence scores using $\textrm{pert}^{gran}(s; g)$ (\autoref{sec:granular}).
We vary $g$ from one to ten (number of bins varies from two to eleven). 
Compared to continuous rewards, for both pairs of languages,
$\Delta$\persentence is not affected with $g$ at least five (at least six bins).
As granularity decreases, $\Delta$\persentence  monotonically degrades.
However, even when $g = 1$ (scores are either 0 or 1), the models still improve by at least a point.

\paragraph{High-variance Rewards.} We simulate noisy rewards using the model of human rating variance 
$\textrm{pert}^{var}(s; \lambda)$ (\autoref{sec:variance}) with 
$\lambda \in \{0.1, 0.2, 0.5, 1, 2, 5 \}$. 
Our models can withstand an amount of about 20\% the variance in our human eval 
data without dropping in $\Delta$\persentence.
When the amount of variance attains 100\%, matching the amount of variance in the human 
data, $\Delta$\persentence go down by 
about 30\% for both pairs of languages. 
As more variance is injected, the models degrade quickly but still improve from
the pre-trained models.
Variance is the most detrimental type of perturbation to NED-A2C 
among the three aspects of human ratings we model. 

\paragraph{Skewed Rewards.} We model skewed raters using $\textrm{pert}^{skew}(s; \rho)$ (\autoref{sec:skew}) with $\rho \in \{0.25, 0.5, 0.67, 1, 1.5, 2, 4\}$.
NED-A2C is robust to skewed scores. 
$\Delta$\persentence is at least 90\% of unskewed scores for most skew values. 
Only when the scores are extremely harsh ($\rho = 4$) does $\Delta$\persentence degrade 
significantly (most dramatically by 35\% on Zh-En). 
At that degree of skew, a score of 0.3 is suppressed to be less than 0.08, giving
little signal for the models to learn from.
On the other spectrum, the models are less sensitive to motivating scores as 
\persentence is unaffected on Zh-En and only decreases by 7\% on De-En.

\subsection{Held-out Translation Quality}
\label{sec:heldout_results}

Our method also improves pre-trained models in \heldout, a metric that
correlates with translation quality better than \persentence
(\autoref{tab:clean_bleu}).  When scores are perturbed by our rating
model, we observe similar patterns as with \persentence: the models
are robust to most perturbations except when scores are very coarse,
or very harsh, or have very high variance (\autoref{fig:noise}, second
row).  Supervised learning improves \heldout better, possibly because
maximizing log-likelihood of reference translations correlates more
strongly with maximizing \heldout of predicted translations than
maximizing \persentence of predicted translations.


\section{Related Work and Discussion} \label{sec:related}

Ratings provided by humans can be used as effective learning signals
for machines.
Reinforcement learning has become the \textit{de facto} standard for
incorporating this feedback across diverse tasks such as robot
voice control~\cite{tenorio2010dynamic}, myoelectric control~\cite{pilarski2011online}, and virtual assistants~\cite{isbell2001social}.
Recently, this learning framework has been combined with recurrent neural networks to solve  
machine translation~\cite{bahdanau2016actor}, dialogue generation~\cite{li2016deep}, neural architecture search~\cite{zoph2016neural}, and device placement~\cite{mirhoseini2017device}. 
Other approaches to more general structured prediction under bandit feedback \cite{daume15lols,sokolov2016learning,sokolov2016stochastic} show the broader efficacy of this framework.
\citet{ranzato2015sequence} describe MIXER for training neural encoder-decoder models, 
which is a reinforcement learning approach closely related to ours but requires a policy-mixing strategy and only uses a linear critic model. 
Among work on bandit MT, ours is closest to \citet{kreutzer17bandit}, 
which also tackle this problem using neural encoder-decoder models, 
but we 
(a) take advantage of a state-of-the-art reinforcement learning method;
(b) devise a strategy to simulate noisy rewards; and
(c) demonstrate the robustness of our method on noisy simulated rewards. 

Our results show that bandit feedback can be an effective
feedback mechanism for neural machine translation systems.
This is \emph{despite} that errors in human annotations hurt machine learning models in many NLP tasks~\cite{snow2008cheap}.
An obvious question is whether we could extend our framework to model individual annotator preferences \cite{passonneau2014benefits} or learn personalized models \cite{mirkin2015motivating,rabinovich2016personalized}, and handle heteroscedastic noise \cite{park1966estimation,kersting2007most,antos2010active}.
Another direction is to apply active learning techniques to reduce the
sample complexity required to improve the systems or to extend to
richer action spaces for problems like simultaneous translation, which requires
prediction~\cite{Grissom:He:Boyd-Graber:Morgan-2014} and
reordering~\cite{He-15} among other strategies to both minimize delay
and effectively translate a sentence~\cite{He-2016}.

\section*{Acknowledgements}

Many thanks to Yvette Graham for her help with the WMT human evaluations data.
We thank UMD CLIP lab members for useful discussions that led to the
ideas of this paper.
We also thank the anonymous reviewers for their thorough and insightful comments.
This work was supported by NSF grants IIS-1320538.
Boyd-Graber is also partially supported by NSF grants IIS-
1409287, IIS-1564275, IIS-IIS-1652666, and NCSE-1422492.
Daum{\'e} III is also supported by NSF grant IIS-1618193,
as well as an Amazon Research Award.
Any opinions, findings, conclusions,
or recommendations expressed here are those of
the authors and do not necessarily reflect the view of the
sponsor(s).

\appendix
\section{Neural MT Architecture}
Our neural machine translation (NMT) model consists of an encoder and a decoder, 
each of which is a recurrent neural network (RNN).
We closely follow \cite{luong2015effective} for the structure of our model.
It directly models the posterior distribution 
$P_{\vec \theta}(\vec y \mid \vec x)$ of translating a source sentence $\vec x = (x_1, \cdots, x_n)$ to
a target sentence $\vec y = (y_1, \cdots, y_m)$:
\begin{align}
  P_{\vec \theta}(\vec y \mid \vec x) = \prod_{t = 1}^m P_{\vec \theta}(y_t \mid \vec y_{<t}, \vec x)
\end{align}
where $\vec y_{<t}$ are all tokens in the target sentence prior to $y_t$. 

Each local disitribution $P_{\vec \theta}(y_t \mid \vec y_{<t}, \vec x)$ is modeled as a multinomial
distribution over the target language's vocabulary. We compute this distribution by applying a linear transformation followed by a softmax function on the decoder's output vector 
$\vec h_t^{dec}$: 
\begin{align}
  P_{\vec \theta}(y_t \mid \vec y_{<t}, \vec x) &= \mathrm{softmax}(\vec W_s \ \vec h_t^{dec}) \\
  \vec h_{t}^{dec} &= \tanh (\vec W_o [\tilde{\vec h}_t^{dec}; \vec c_t]) \\
  \vec c_t &= \mathrm{attend}(\tilde{\vec h}_{1:n}^{enc}, \tilde{\vec h}_t^{dec})
\label{eqn:softmax}
\end{align} where 
$[\cdot;\cdot]$ is the concatenation of two vectors, 
$\mathrm{attend}(\cdot,\cdot)$ is an attention mechanism,
$\tilde{\vec h}_{1:n}^{enc}$ are all encoder's hidden vectors and 
$\tilde{\vec h}_t^{dec}$ is the decoder's hidden vector at time step $t$.
We use the ``general'' global attention in~\cite{luong2015effective}.

During training, the encoder first encodes $\vec x$ to a continuous
vector $\Phi(\vec x)$, which is used as the initial hidden vector for the decoder. 
In our paper, $\Phi(\vec x)$ simply returns the last hidden vector of the encoder.
The decoder performs RNN updates to produce a sequence of hidden 
vectors:
\begin{equation}
\begin{split}
  \tilde{\vec h}_0^{dec} &= \Phi(\vec x) \\
  \tilde{\vec h}_t^{dec} &= f_{\vec \theta} 
  \left(\tilde{\vec h}_{t - 1}^{dec}, \left[ \vec h_{t-1}^{dec}; e(y_t) \right] \right)
\end{split}
\label{eqn:decoder}
\end{equation} where $e(.)$ is a word embedding lookup function and $y_t$ is the 
ground-truth token at time step $t$. 
Feeding the output vector $\vec h_{t - 1}^{dec}$ to the next step is known as ``input feeding''.

At prediction time, the ground-truth token $y_t$ in \autoref{eqn:decoder} 
is replaced by the model's own prediction $\hat{y}_t$:
\begin{equation}
  \hat{y}_t = \arg \max_y P_{\vec \theta}(y \mid \hat{\vec y}_{<t}, \vec x)
\end{equation}

In a supervised learning framework, an NMT model is typically trained under the 
maximum log-likelihood objective:
\begin{equation}
\begin{split}
  \max_{\vec \theta} \mathcal{L}_{sup}(\vec \theta) &= 
  \max_{\vec \theta} \mathbb{E}_{(\vec x, \vec y) \sim D_{\textrm{tr}}} 
  \left[ \log P_{\vec \theta} \left( \vec y \mid \vec x \right) \right]
\end{split}
\end{equation} where $D_{\textrm{tr}}$ is the training set.  
However, this learning framework is not applicable to bandit learning since 
ground-truth translations are not available.

\bibliography{emnlp2017}
\bibliographystyle{emnlp_natbib}

\end{document}